\pdfoutput=1

\documentclass[11pt]{article}

\usepackage[final]{acl}

\usepackage{times}
\usepackage{latexsym}
\usepackage{comment}
\usepackage[T1]{fontenc}

\usepackage[utf8]{inputenc}

\usepackage{microtype}

\usepackage{inconsolata}

\usepackage{graphicx}
\usepackage{amsmath}
\usepackage{enumitem}
\usepackage{multirow}
\usepackage{multicol}
\usepackage{booktabs}
\usepackage{arydshln}
\usepackage{color}
\usepackage{makecell}
\usepackage{subcaption}
\usepackage{lipsum}

\usepackage{colortbl}
\definecolor{mygray}{gray}{.9}

\usepackage{listings}
\usepackage{xcolor}

\usepackage{tcolorbox}

\definecolor{light-bg}{RGB}{250,250,250}
\definecolor{light-keyword}{RGB}{0,119,170}
\definecolor{light-string}{RGB}{221,74,104}
\definecolor{light-comment}{RGB}{112,128,144}
\definecolor{light-frame}{RGB}{240,240,240}

\title{
    \includegraphics[height=1.5em]{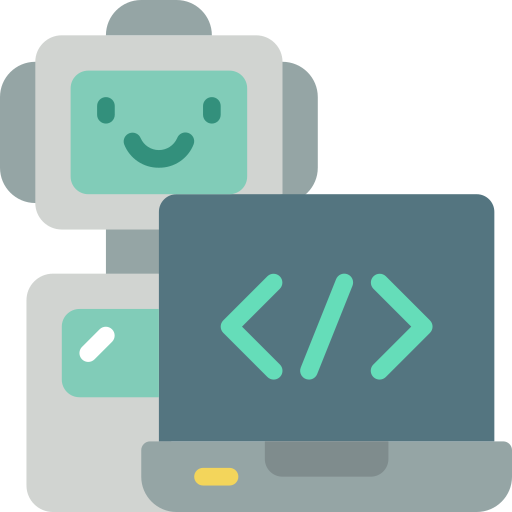} 
    \texttt{ToolCoder}: A Systematic Code-Empowered Tool Learning Framework \\ 
    for Large Language Models
}
\author{Hanxing Ding$^{1,2}$\thanks{\ \ Equal contributions.}\quad
Shuchang Tao$^{3}$\footnotemark[1] \quad
Liang Pang$^{1}$\thanks{\ \ Corresponding author}\quad
Zihao Wei$^{1,2}$\quad\\
\textbf{Jinyang Gao}$^{3}$\quad
\textbf{Bolin Ding}$^{3}$\quad
\textbf{Huawei Shen}$^{1,2}$
\textbf{Xueqi Cheng}$^{1,2}$\\
 $^{1}$State Key Laboratory of AI Safety, Institute of Computing Technology, CAS \\
 $^{2}$ University of Chinese Academy of Sciences \quad $^{3}$ Alibaba Group \\
 \texttt{\{dinghanxing18s, pangliang, weizihao22z, shenhuawei, cxq\}@ict.ac.cn} \\
}

\begin{document}
\maketitle
\begin{abstract}
Tool learning has emerged as a crucial capability for large language models (LLMs) to solve complex real-world tasks through interaction with external tools. Existing approaches face significant challenges, including reliance on hand-crafted prompts, difficulty in multi-step planning, and lack of precise error diagnosis and reflection mechanisms. We propose \texttt{ToolCoder}, a novel framework that reformulates tool learning as a code generation task. Inspired by software engineering principles, \texttt{ToolCoder} transforms natural language queries into structured Python function scaffold and systematically breaks down tasks with descriptive comments, enabling LLMs to leverage coding paradigms for complex reasoning and planning. It then generates and executes function implementations to obtain final responses. Additionally, \texttt{ToolCoder} stores successfully executed functions in a repository to promote code reuse, while leveraging error traceback mechanisms for systematic debugging, optimizing both execution efficiency and robustness. Experiments demonstrate that \texttt{ToolCoder} achieves superior performance in task completion accuracy and execution reliability compared to existing approaches, establishing the effectiveness of code-centric approaches in tool learning. Our code is available at \href{https://github.com/dhx20150812/ToolCoder}{\color{red}{this link}}.
%
\end{abstract}
\section{Introduction}
As large language models (LLMs) continue to advance, tool learning has emerged as a critical capability, enabling LLMs to solve complex real-world tasks through dynamic interaction with external tools and APIs~\cite{DBLP:journals/corr/abs-2405-17935}. This capability extends LLMs' problem-solving abilities and allows them to adapt to dynamic environments and specialized domains~\cite{DBLP:conf/iclr/00020LCYPJ24}.

\begin{figure}[t]
    \centering
    \includegraphics[width=\columnwidth]{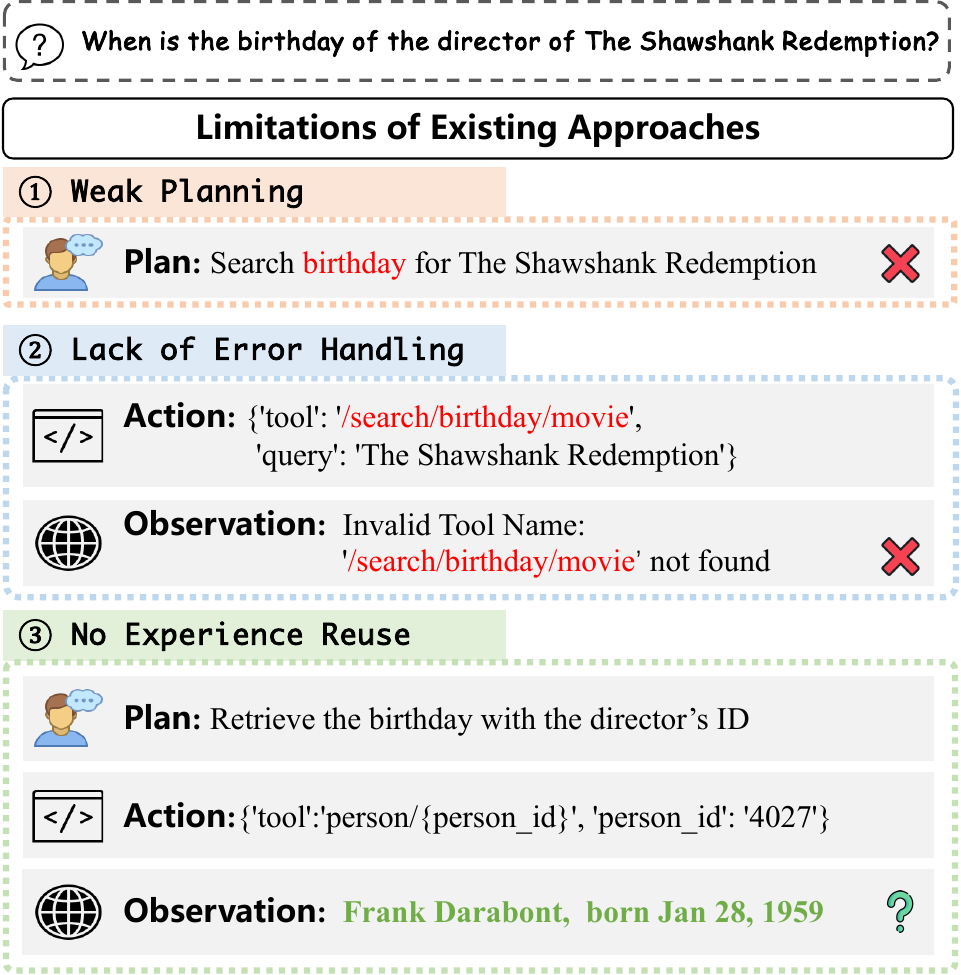}
    \caption{Illustration of key limitations in existing approaches: weak planning strategies, insufficient error handling, and lack of experience reuse.}
    \label{fig:intro}
    \vspace{-0.6cm}
\end{figure}

Current approaches for tool learning predominantly follow a sequential \textit{plan-execute-observe} paradigm~\cite{DBLP:conf/iclr/YaoZYDSN023,DBLP:journals/corr/abs-2303-09014,DBLP:conf/nips/LuPCGCWZG23}, where LLMs iteratively plan actions, execute tools, and observe outcomes. 
While frameworks like ReAct~\cite{DBLP:conf/iclr/YaoZYDSN023} and Chameleon~\cite{DBLP:conf/nips/LuPCGCWZG23} have demonstrated success in simple task handling, they suffer from several limitations. 
As shown in Figure~\ref{fig:intro}, these methods rely heavily on hand-crafted prompts and natural language reasoning, making multi-step planning difficult and often leading to unreliable performance in complex tasks~\cite{DBLP:journals/corr/abs-2407-03007,DBLP:conf/icml/WangCY0L0J24}.
Moreover, they lack precise error diagnosis and reflection mechanisms—when execution fails, they are unable to identify the actual causes of errors or propose targeted corrective actions~\cite{DBLP:conf/icml/WangCY0L0J24,wang-etal-2024-llms-imaginarium}. 
Furthermore, current frameworks are unable to accumulate and reuse successful experiences from past executions, treating each query in isolation and repeatedly solving similar problems from scratch~\cite{DBLP:conf/iclr/YuanC000J24}.
These limitations hinder the robustness, adaptability, and scalability of tool-learning systems.

In this paper, we propose \texttt{ToolCoder}, a novel code-empowered framework that re-formulates tool learning as a code generation task. 
Inspired by software engineering principles, \texttt{ToolCoder} systematically addresses these challenges in tool learning through these key stages: 
First, \texttt{ToolCoder} converts vague natural language queries into structured Python function scaffold, analogous to the requirement analysis in software engineering, enabling clearly defined input-output specifications. 
Then, following modular design principles, \texttt{ToolCoder} systematically breaks down scaffold into subtasks with descriptive comments, enabling LLMs to leverage coding paradigms for complex reasoning and planning.
In the implementation phase, \texttt{ToolCoder} generates well-structured runnable code for both sub-function and main functions, then executes them to obtain final responses. Successfully executed code snippets are stored in a function repository to accumulate proven implementations, helping solving similar problems.
For failed executions, \texttt{ToolCoder} leverages explicit Python's error traceback mechanism for accurate error diagnosis, significantly improving the reliability of tool learning system.

To evaluate \texttt{ToolCoder}'s effectiveness, we conduct comprehensive experiments on multiple benchmark datasets~\cite{DBLP:journals/corr/abs-2306-06624,li-etal-2023-api}. Our experimental results demonstrate that \texttt{ToolCoder} significantly outperforms SOTA approaches in all metrics. We also validate the necessity of each proposed component. Furthermore, we observe that \texttt{ToolCoder} brings more substantial improvements in code LLMs compared to base LLMs. These findings confirm the superiority of systematic code-centric framework and its potential for advancing tool learning systems.

The main contributions of this work are:
\begin{itemize}[leftmargin=0.6cm, itemindent=0cm, itemsep=0pt]
    \item[$\bullet$] We propose \texttt{ToolCoder} framework that re-formulates tool learning as a code generation task, leveraging both software engineering principles and LLMs' code generation capabilities.
    \item[$\bullet$] We design a systematic architecture that converts natural language task into modular code components, implements and executes them to generate final response. We also incorporate error diagnosis and code reuse mechanisms for efficiency and reliability.
    \item[$\bullet$] Extensive experiments on benchmarks demonstrate \texttt{ToolCoder} significantly improve the success rate and reliability of tool learning.
\end{itemize}
\section{Related works}
\subsection{Text-based Tool Learning}
Recent approaches have focused on enhancing LLMs by integrating external tools to improve their problem-solving capabilities~\cite{DBLP:journals/corr/abs-2404-11584,DBLP:journals/corr/abs-2405-17935,qian-etal-2023-creator,DBLP:conf/iclr/YuanC000J24}. Typically, these methods follow a sequential \textit{plan-execute-observe} paradigm and involve utilizing text few-shot prompt to decompose complex tasks into sequential steps, with the LLM executes each tool individually and incorporates the results into its context to manage data flow dependencies~\cite{DBLP:journals/corr/abs-2402-18157,DBLP:conf/nips/ShinnCGNY23,shi-etal-2024-learning,DBLP:journals/corr/abs-2303-09014}. For instance, ReAct~\cite{DBLP:conf/iclr/YaoZYDSN023} enables LLMs to generate both reasoning traces and action plans, facilitating more effective task execution. Similarly, Chameleon~\cite{DBLP:conf/nips/LuPCGCWZG23} employs a compositional reasoning framework that dynamically assembles specialized tools to address complex reasoning tasks. RestGPT~\cite{DBLP:journals/corr/abs-2306-06624} introduces a coarse-to-fine online planning approach, allowing LLMs to iteratively refine task decomposition. However, these text-based methods heavily rely on carefully crafted prompts and struggle with complex multi-step tasks, limiting reliability. \texttt{ToolCoder} addresses these challenges by re-formulating tool learning as code generation, leveraging structured code generation and software engineering principles for better adaptability.


\begin{figure*}[t]
    \centering
    \includegraphics[width=2\columnwidth]{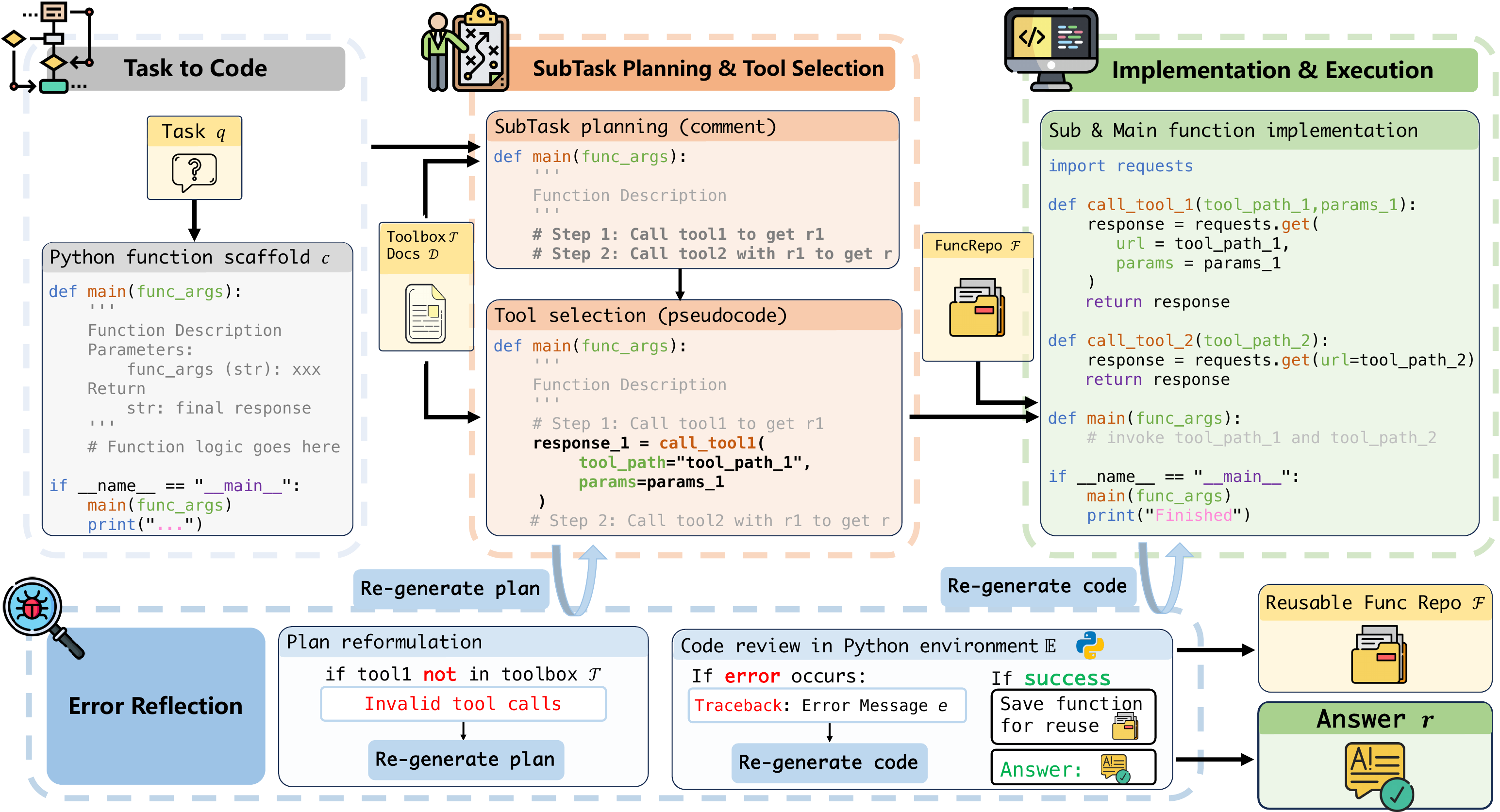}
    \caption{Overview of our \texttt{ToolCoder} framework, which converts a user task into executable Python code through task-to-code transformation, tool planning and selection, code implementation, and error reflection. The framework also leverages reusable function snippets to enhance efficiency and accuracy.}
    \label{fig:overview}
    \vspace{-0.5cm}
\end{figure*}

\subsection{Code-based Tool Learning}
Code pre-training has been shown to significantly enhance the chain-of-thought (CoT) performance of LLMs, enabling them to logically decompose complex tasks into smaller, more manageable sub-tasks~\cite{lyu-etal-2023-faithful,DBLP:conf/iclr/ChengX0LNHXROZS23,DBLP:conf/sigir/YeHYLHL23}. Recent studies have further highlighted the potential of code-empowered LLMs to improve the planning and reasoning capabilities of these models~\cite{DBLP:journals/corr/abs-2401-00812,DBLP:journals/tmlr/ChenM0C23,DBLP:conf/icml/GaoMZ00YCN23}. For example, certain approaches enable LLMs to generate programmatic chains of thought to solve complex numeric reasoning tasks, yielding impressive performance~\cite{DBLP:journals/tmlr/ChenM0C23,DBLP:conf/icml/GaoMZ00YCN23}. In the context of tool learning, LLMs can generate code snippets as actions~\cite{DBLP:conf/icml/WangCY0L0J24}, leveraging widely used Python functions and simplifying operations such as lengthy for-loops~\cite{DBLP:journals/corr/abs-2405-16533}.
Previous approaches generate Python code in a single pass, lacking iterative refinement, intent alignment, and learning from past executions~\cite{DBLP:journals/corr/abs-2308-00675,DBLP:journals/corr/abs-2401-06201,DBLP:conf/icml/WangCY0L0J24}. Our \texttt{ToolCoder} framework addresses these limitations by integrating software engineering principles, including modular code construction, stepwise verification, and experience-driven refinement, ensuring the generated code is executable and robust.

\section{Methodology}
We formulate the tool learning task, present our motivation, and propose the \texttt{ToolCoder} framework.

\subsection{Task Formulation and Motivation}
The goal of tool learning is to enable a LLM $\mathcal{M}$ to accomplish natural language tasks by generating and executing appropriate tool sequences. Given a task $q$, the LLM selects appropriate tools from a toolbox $\mathcal{T} = \{t_1, t_2, \ldots, t_{|\mathcal{T}|}\}$, where each tool $t_i$ is accompanied by its documentation $d_i \in \mathcal{D}$ describing its functionality, arguments, and output schema. Through tool interaction, $\mathcal{M}$ generates a solution response $r$ that finally addresses the task.

Existing methods struggle with complex task planning and robust execution. To address these limitations, We draw inspiration from software engineering principles, which emphasize modular design, error diagnosis, and code reusability. Furthermore, by expressing complex tasks through structured code, LLMs can leverage established programming paradigms to enhance their systematic reasoning and planning capabilities~\cite{DBLP:conf/icml/GaoMZ00YCN23, DBLP:conf/icml/WangCY0L0J24}.

Inspired by these insights, we reformulate tool learning as a code generation task. This reformulation enables us to \emph{apply fundamental software engineering principles throughout the tool learning process: from initial requirement analysis and task decomposition, through modular implementation, to systematic error handling and code reuse}. 
By structuring tool operations as well-defined programming components, we can leverage both the systematic methodology of software engineering and the code generation capabilities of LLMs.

\subsection{\texttt{ToolCoder}: Code-Empowered Tool Learning Framework}
We propose a novel framework \texttt{ToolCoder}, which converts queries into Python function scaffolds (\S~\ref{sec:task_to_code}), performs subtask planning and tool selection (\S~\ref{sec:planning}), and generates executable implementations (\S~\ref{sec:main_function}) with error reflection mechanisms (\S~\ref{sec:reflection}), as illustrated in Figure~\ref{fig:overview}.

\subsubsection{Task-to-Code Transformation}
\label{sec:task_to_code}

Our \texttt{ToolCoder} starts by identifying user needs and transforming the natural language task $q$ into a structured Python function scaffold $c$, aligning with the requirements analysis phase in software engineering. This process can be formalized as:
\begin{equation}
\textstyle
\setlength\abovedisplayskip{0.2cm}
\setlength\belowdisplayskip{0.2cm}
\begin{aligned}
c = \mathcal{M}_{T2C}(q),
\end{aligned}
\end{equation}
where $\mathcal{M}_{T2C}$ represents the task-to-code transformation module, following the prompt template shown in Appendix Figure~\ref{fig:task_to_code_prompt}. The scaffold $c$, as shown in Figure~\ref{fig:scaffold}, includes several key components: a descriptive function name that indicates the intended task, a parameter list that specifies required inputs, and a comprehensive docstring that documents the function's purpose, parameter descriptions, and expected return value. The function body is deliberately left empty at this stage, serving as a structured placeholder to be filled during subsequent planning and tool execution phases.

\begin{figure}[t]
\centering
\begin{lstlisting}[language=python, basicstyle=\ttfamily\tiny]
def get_directed_movie_count(director_name: str) -> int:
    '''
    Get the number of movies directed by a specific director.

    Parameters:
        director_name (str): The full name of the director whose movies to count.

    Returns:
        int: The number of movies directed by the specified director.
    '''
    # Function logic goes here

if __name__ == "__main__":
    number_of_movies_directed = get_directed_movie_count(director_name="Sofia Coppola")
    print("Number of movies directed by Sofia Coppola:", number_of_movies_directed)
\end{lstlisting}
\caption{An example of the Python function scaffold $c$ generated by \texttt{ToolCoder} for the query: \textit{give me the number of movies directed by Sofia Coppola}.}
\label{fig:scaffold}
\vspace{-0.3cm}
\end{figure}

By converting natural language queries into structured code scaffolds, \texttt{ToolCoder} clearly understands the requirements, precise input and output specifications, and clear task goals, laying a solid foundation for task decomposition.

\subsubsection{Subtask Planning and Tool Selection}
\label{sec:planning}
This stage focuses on subtask planning and tool selection, decomposing tasks into modular components and determining appropriate tools, following modular design in software engineering. Given candidate toolbox $\mathcal{T}$ and the generated Python scaffold $c$, \texttt{ToolCoder} translates high-level tasks into concrete, actionable subtasks:
\begin{equation}
\textstyle
\setlength\abovedisplayskip{0.2cm}
\setlength\belowdisplayskip{0.2cm}
\begin{aligned}
\boldsymbol{s} = \mathcal{M}_{TP}(c, \mathcal{T}) = \{s_1, s_2, \dots, s_m\},
\end{aligned}
\end{equation} 
where $\mathcal{M}_{TP}$ is the subtask planning module  (implemented with the prompt in Appendix Figure~\ref{fig:subtask_planning_prompt}). $\mathcal{M}_{TP}$ analyzes the Python scaffold $c$ and generates concrete subtasks $\boldsymbol{s} = \{s_1, s_2, \dots, s_m\}$. These subtasks are then embedded as structured code comments within scaffold $c$, forming a subtask-based plan $c_{\boldsymbol{s}}$ that outlines all the necessary steps.

The tool selection module $\mathcal{M}_{TS}$ (prompt in Appendix Figure~\ref{fig:tool_selection_prompt}) systematically analyzes the subtask-based plan $c_{\boldsymbol{s}}$ and references the documentation of each tool to understand their input-output specifications.  
Based on the analysis, \texttt{ToolCoder} generates pseudocode $c_{\boldsymbol{p}}$ whose main function body contains proper tool invocation sequences and clear data flow pathways, such as parsing the necessary parameters from the previous API to input the next API call. Each necessary tool call is placed with a sub-function placeholder, whose specific functionality is not implemented.

Through structured planning and tool selection, \texttt{ToolCoder} maximizes the model's reasoning potential by leveraging coding paradigms, ensuring reliable and systematic task completion.

\subsubsection{Implementation and Execution}
\label{sec:main_function}
\paragraph{Implementation}
In this stage, \texttt{ToolCoder} implements the specific functionality of sub-function placeholder, using the API documentation of selected tools $\mathcal{T}_{\boldsymbol{s}}$ and a repository of correctly-executed subfunction $\mathcal{F}$, assembling them into an executable program $F$.
\begin{equation}
\textstyle
\setlength\abovedisplayskip{0.2cm}
\setlength\belowdisplayskip{0.2cm}
\begin{aligned}
F = \mathcal{M}_{CG}(c_{\boldsymbol{p}}, \mathcal{T}_{\boldsymbol{s}}, \mathcal{F}),  
\end{aligned}
\end{equation} 
where $\mathcal{M}_{CG}$ is the code generation module (prompt in Appendix Figure~\ref{fig:main_function_template}), $\mathcal{T}_{\boldsymbol{s}}$ denotes the subset of candidate tools selected based on the pseudocode $c_{\boldsymbol{p}}$, and $\mathcal{F}$ represents the reusable repository of successfully executed functions. 
For each subplan $s_i$, \texttt{ToolCoder} implements corresponding sub-functions by leveraging tool documentation from $\mathcal{T}_{\boldsymbol{s}}$ and reusable repository $\mathcal{F}$.

\paragraph{Execution}
Once the function $F$ is generated, it is executed within the Python environment $\mathbf{E}$ to produce response $r$ or exceptions $e$:  
\begin{equation}
\textstyle
\setlength\abovedisplayskip{0.2cm}
\setlength\belowdisplayskip{0.2cm}
\begin{aligned}
r, e = \mathbf{E}(F).
\end{aligned}
\end{equation}
If program $F$ is successfully executed, the generated response $r$ will be returned as the final answer to the user’s task $q$. Otherwise, the error message $e$ triggers an error reflection process to identify and resolve any issues.

\paragraph{Reusable Function Repository}
To facilitate the implementation process, \texttt{ToolCoder} maintains a reusable function repository $\mathcal{F}$, which stores successfully executed sub-function snippets that implement specific API functionalities. During task execution, \texttt{ToolCoder} extracts these successful sub-functions from program $F$ and adds them to repository $\mathcal{F}$, creating a collection of reusable implementations with their execution contexts.

Through collecting these sub-function snippets, \texttt{ToolCoder} builds a reliable function repository for each tool used. This enables efficient code reuse in future tasks, preventing potential errors and eliminating redundant development while enhancing system efficiency and reliability.

\subsubsection{Error Reflection}
\label{sec:reflection}
To mitigate common errors like vague planning, invalid tool selection, and execution failures, \texttt{ToolCoder} leverages explicit exception tracebacks to identify and correct issues, and iteratively refine task plans and enhance the reliability of the system.

\paragraph{Plan Reformulation}  
If the plan $c_{\boldsymbol{s}}$ generated by the task decomposition module $\mathcal{M}_{TD}$ includes invalid or non-existent tools, \texttt{ToolCoder} invokes planning reformulation strategy to refine the plan. Specifically, it detects errors by cross-referencing each tool in the plan against the available tools in $\mathcal{T}$. For every invalid tool, it generates explicit instructions for the LLM, guiding LLM to select an appropriate alternative from $\mathcal{T}$ to fulfill the required functionality. This ensures that the reformulated plan $c_{\boldsymbol{s}^{'}}$ consists entirely of tools that are present in the candidate toolbox.

\paragraph{Code Review}
If the initial function $F$ fails, producing an error $e$, \texttt{ToolCoder} employs  code review to analyze the error, identify its cause, and generate corrections. Python's clear and detailed exception tracebacks allow precise pinpointing of the error location and facilitate understanding of the underlying issue. The function $F$ is revised based on the current error $e$ and then re-executed using $\mathbf{E}$, yielding a new result $r$ and an updated error state $e$. This process is performed iteratively, typically up to three times, until execution succeeds (i.e., $e = \emptyset$) or the maximum number of attempts is reached.
\section{Experiments}
In this section, we describe the experimental setup, including the datasets, evaluation metrics, and baseline methods. We then present the main experimental results, followed by a comprehensive analysis that offers an in-depth examination of the model's capabilities and efficiency.

\begin{table*}[t]
\centering
\resizebox{\linewidth}{!}{
\begin{tabular}{lccccccc}
\toprule
\multirow{2}{*}{\textbf{Methods}} & \multicolumn{3}{c}{\textbf{RestBench-TMDB}} & \multicolumn{2}{c}{\textbf{RestBench-Spotify}}  & \textbf{API-Bank (LV1)} & \textbf{API-Bank (LV2)} \\
\cline{2-4}\cline{5-6}\cline{6-7}\cline{7-8}
& Success(\%) & Accuracy(\%)  & Path(\%) & \multicolumn{1}{c}{Success(\%)} & \multicolumn{1}{c}{Path(\%)} & Correctness(\%)     & Correctness(\%)  \\
\midrule
\rowcolor{mygray}
\textit{Text-based Tool Learning}  &    &   &         &        &      &   & \\
\midrule
ReAct~\cite{DBLP:conf/iclr/YaoZYDSN023}  & 76.0   & 48.0   & 50.0    & 68.42  &  52.63   & 73.93  & 56.30   \\
Chameleon~\cite{DBLP:conf/nips/LuPCGCWZG23}    & 75.0    & 45.0     & 52.0  &  70.18  &  63.16  & 74.87   & 37.04  \\
ConAgents~\cite{shi-etal-2024-learning}    &   72.0  &   38.0    &   57.0   &  64.92  &   68.42   &  67.26 &   36.24     \\
RestGPT~\cite{DBLP:journals/corr/abs-2306-06624}    & 83.0    & 33.0     & 49.0      &   61.41   &   57.89  &  65.47  &  34.83    \\
EasyTool~\cite{DBLP:journals/corr/abs-2401-06201}       &  75.0       &  45.0     & 62.0     &  62.19  &   64.92   &  74.97 &   58.24  \\
\midrule
\rowcolor{mygray}
\textit{Code-based Tool Learning}  &    &   &         &        &      &   & \\
\midrule
ATC~\cite{DBLP:journals/corr/abs-2405-16533} &   78.0  &   58.0    &   62.0   &  65.47  &   68.42   &  70.21  &   52.18     \\
CodeAct~\cite{DBLP:conf/icml/WangCY0L0J24}     &  80.0  &  56.0 &   67.0  & 71.93  &   66.67   & 75.94   & 54.07     \\
\midrule
\texttt{ToolCoder}    &  \textbf{85.0}  &  \textbf{78.0}   &  \textbf{83.0}  &  \textbf{87.72}   &   \textbf{78.95}  &  \textbf{83.08}  &  \textbf{62.41} \\
\quad \textit{w/o} Reusable Repository  & 83.0   & 71.0  & 78.0   &    78.95    &  71.93    &  83.08$^{*}$ &  62.41$^{*}$ \\
\quad \textit{w/o} Error Reflection  &  75.0  & 65.0  & 77.0    &    73.68    &  70.18   &  79.85  &  58.02 \\
\bottomrule
\end{tabular}}
\caption{Main experimental results on RestBench and API-Bank datasets. All methods are implemented with \textit{gpt-4o-mini}. The best results are bolded. $^*$ Results from the API-Bank dataset are reported without the reusability mechanism due to its built-in API calls in this dataset.}
\label{tab:main}
\vspace{-0.3cm}
\end{table*}

\subsection{Experimental Settings}
\paragraph{Datasets.} To validate the effectiveness of our method in tool learning, we conduct experiments on two well-established benchmarks: RestBench~\cite{DBLP:journals/corr/abs-2306-06624} and API-Bank~\cite{li-etal-2023-api}. RestBench comprises two real-world scenarios: TMDB, a human-annotated dataset with 100 tasks utilizing 54 tools, designed for movie-related scenarios; and Spotify, which features 40 tools tailored to music-related tasks. Each tool in RestBench is paired with a detailed RESTful API document, making it inherently suitable for benchmarking the real-world tool-learning capabilities of LLMs. API-Bank consists of 73 commonly used APIs and 314 tool-use dialogues with 753 manually annotated API calls, designed to evaluate the effectiveness of LLMs in utilizing tools.

\paragraph{Evaluation metrics.} For RestBench~\cite{DBLP:journals/corr/abs-2306-06624}, we adopt their evaluation framework and utilize three metrics: success rate (\textbf{Success\%}), which relies on human evaluation to determine whether the model’s output effectively fulfills the user query; correct path rate (\textbf{Path\%}), which measures the proportion of ground-truth tools correctly included in the model-generated tool calls. We also introduce \textbf{Accuracy\%}, specifically for the TMDB scenario, to assess whether the generated answers align with the ground-truth answers. These ground-truth answers are derived by executing the annotated solution paths, providing a reliable benchmark for comparison. For API-Bank~\cite{li-etal-2023-api}, we evaluate API-calling performance using their \textbf{Correctness\%} metric, which compares the ground-truth API outputs with the execution results of the model-generated API calls.

\subsection{Baselines} We compare our method against seven well-known baselines: (1) ReAct~\cite{DBLP:conf/iclr/YaoZYDSN023}, prompting LLMs to generate interleaved chains of thought and actions. (2) Chameleon~\cite{DBLP:conf/nips/LuPCGCWZG23}, an LLM-based agent that creates multi-step plans for tool usage and executes them sequentially. (3) ConAgents~\cite{shi-etal-2024-learning}, which facilitates the collaboration of three specialized LLMs to solve complex tasks. (4) RestGPT~\cite{DBLP:journals/corr/abs-2306-06624} features a coarse-to-fine planning module and a tool executor. (5) EasyTool~\cite{DBLP:journals/corr/abs-2401-06201} converts lengthy and diverse tool documentation into unified, concise instructions for easier tool usage. (6) ATC~\cite{DBLP:journals/corr/abs-2405-16533} utilizes a chain of tools through programming and proposes a black-box probing method. (7) CodeAct~\cite{DBLP:conf/icml/WangCY0L0J24}, enabling LLMs to generate executable code snippets as actions to interact with tools.

\subsection{Main Results}
The experimental results in Table~\ref{tab:main} provide a comprehensive evaluation of \texttt{ToolCoder} in comparison to existing text-based and code-based tool learning approaches. These results highlight the significant performance improvements achieved by \texttt{ToolCoder}, validating the effectiveness of its code-empowered framework.

From the experimental results presented in Table~\ref{tab:main}, text-based approaches demonstrate limited effectiveness in complex tool learning scenarios, as evidenced by their low path rates and accuracy scores. This underperformance stems from two key limitations: the inherent constraints of natural language prompting for structured reasoning, and their inability to effectively handle execution errors. While code-based methods like ATC and CodeAct show significant improvements in path planning, achieving higher success rates than their text-based counterparts, they still face challenges in maintaining consistent performance across different scenarios. This is particularly evident in their accuracy scores, suggesting that even with better planning capabilities, these methods still lack robust mechanisms for error handling and output verification. 

Building on the strengths of code-empowered LLMs, \texttt{ToolCoder} achieves SOTA performance across all benchmarks compared to baseline methods, consistently achieving the highest success, accuracy, and correct path rates, showcasing its robustness and adaptability. For instance, on RestBench-Spotify, \texttt{ToolCoder} improves the success rate by 10.79\% and the accuracy by 22.22\% compared to the strongest baseline, CodeAct. Similarly, on API-Bank (LV2), it improves correctness by 8.34\% over CodeAct, highlighting its ability to handle multi-step tool interactions. The performance gains stem from \texttt{ToolCoder}'s code-empowered framework, which leverages the reasoning capabilities of code-empowered LLMs and the advantages of systematic software development principles to generate precise tool execution paths and accurate final responses.

\begin{figure}[t]
    \centering
    \includegraphics[width=0.9\columnwidth]{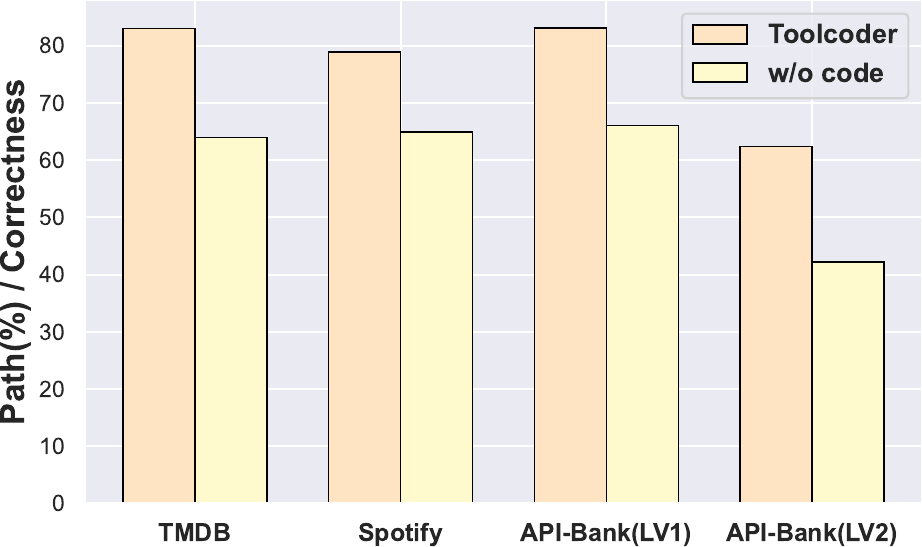}
    \caption{Impact of code-empowered planning on path correctness. Removing the structured Python function scaffold $c$ leads to significant performance degradation across all benchmarks.}
    \label{fig:planning_comparison}
    \vspace{-0.65cm}
\end{figure}

\subsection{Ablation Study}
\paragraph{Effect of Code-Empowered Planning}  
To investigate the effect of code-empowered planning, we remove the structured Python function scaffold $c$ (denoted as \textit{w/o code}) during the tool planning phase and compare its performance with the original \texttt{ToolCoder}. As shown in Figure~\ref{fig:planning_comparison}, the correct path rate drops significantly without $c$ across all benchmarks. This decline highlights the scaffold's role in enhancing the model's reasoning capabilities. Without the scaffold, the model struggles to fully leverage its code-enhanced reasoning abilities, leading to a marked decrease in correct path rate. The absence of $c$ results in a reduced understanding of task objectives and an impaired ability to structure information into coherent plans. These findings underscore the critical importance of the structured Python function scaffold $c$ in activating the model's reasoning potential and maintaining effective task planning.

\begin{figure}[t]
    \centering
    \includegraphics[width=\columnwidth]{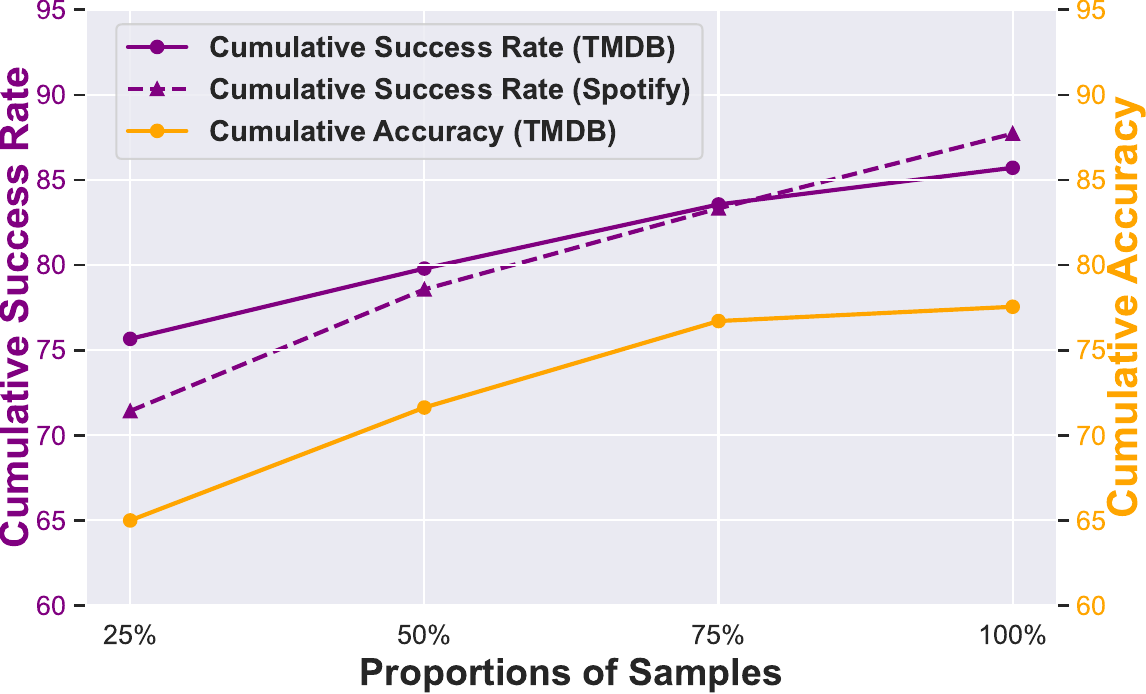}
    \caption{Effect of the function reusability mechanism on cumulative success rate and accuracy across different sample proportions (inference process progresses).}
    \label{fig:cumulative_success_rate}
    \vspace{-0.5cm}
\end{figure}

\paragraph{Effect of Reusable Function Repository}
To assess the impact of the reusable function repository $\mathcal{F}$, we conduct an ablation study by removing the reusability mechanism. As shown in Table~\ref{tab:main}, this results in a 2–5\% performance drop across various benchmarks, highlighting the effectiveness of reusable code snippets.

Furthermore, to gain deeper insights into how reusability enhances tool execution, we analyze the cumulative success rate and cumulative accuracy across different sample proportions, as illustrated in Figure~\ref{fig:cumulative_success_rate}. The results demonstrate a significant performance boost due to the reusability mechanism. As the inference process progresses (with increasing sample proportions on the x-axis), the cumulative success rate steadily grows. This indicates that the model accumulates experience from past successes, aiding in the generation of more robust code. For example, on the TMDB dataset, the cumulative success rate approaches 90\% as more samples are utilized, reflecting improved generalization and execution efficiency. Additionally, the cumulative accuracy on TMDB consistently improves, further confirming that the reusability mechanism not only enhances success rates but also improves answer precision.

\paragraph{Effect of Error Reflection Mechanism}
To assess the impact of the error reflection mechanism, we conduct an ablation study by removing it from \texttt{ToolCoder} and evaluating the resulting performance changes. The results, as shown in Table~\ref{tab:main}, indicate a significant performance drop across multiple datasets when the error reflection mechanism is removed. Both correct path rate and success rates decrease markedly. This decline is attributed to the error reflection mechanism's ability to diagnose planning and execution errors, allowing for iterative refinement and adaptive self-correction. Without this mechanism, the model's reliability is severely compromised. This ablation study demonstrates the critical role of the error reflection mechanism in ensuring robust tool learning and effective adaptation to complex execution challenges.

\begin{table}[t]
\centering
\resizebox{\linewidth}{!}{
\begin{tabular}{lccc}
\toprule
\multirow{2}{*}{\textbf{Methods}} &  \multicolumn{3}{c}{\textbf{ToolAlpaca (Real-world)}} \\
\cmidrule(lr){2-4}
& Procedure & Response & Overall \\ 
\midrule
ReAct & 64.86 & 60.81 & 54.05 \\ 
Chameleon & 68.06 & 58.33 & 57.92 \\ 
CodeAct  & 68.92 & 58.11 & 56.94 \\
\midrule
\texttt{ToolCoder} & \textbf{78.38} & \textbf{75.68} & \textbf{72.97} \\
\quad \textit{w/o} Error Reflection & 76.46 & 68.11 & 67.94 \\
\bottomrule
\end{tabular}}
\caption{Performance comparison of different models on real-world API tasks using the ToolAlpaca dataset. Metrics include Procedure, Response, and Overall scores. All methods are implemented with \textit{gpt-4o-mini}. Due to the small number of questions under each tool category, we do not use the reusability mechanism on this dataset.}
\label{tab:toolalpaca}
\vspace{-0.3cm}
\end{table}

\subsection{Analysis of Generalization Capabilities}
\paragraph{Experiments on New Dataset ToolAlpaca-Real}
To evaluate the generalization performance of \texttt{ToolCoder}, we conduct experiments on the ToolAlpaca dataset~\cite{DBLP:journals/corr/abs-2306-05301} with real-world API tasks. The evaluation framework contains three metrics: Procedure, Response, and Overall. Procedure assesses the model's ability to select appropriate actions, use correct parameters, and execute steps without redundancy. Response measures the model's capability to produce outputs that fully satisfy the user's requirements. Overall is a comprehensive metric that reflects the accuracy and effectiveness of the entire process, encompassing both procedural correctness and response quality.

The experimental results in Table~\ref{tab:toolalpaca} demonstrate that \texttt{ToolCoder} outperforms traditional methods, including ReAct, Chameleon, and CodeAct, across all metrics. \texttt{ToolCoder} achieves significant improvements in Procedure and Response, which lead to a superior Overall performance. These advancements can be attributed to the code-empowered LLM's superior planning and execution capabilities, enabling more accurate action selection and task completion. Additionally, the integration of a reflection strategy further enhances the quality of the responses, resulting in more effective handling of real-world API tasks compared to baselines.

\begin{table}[t]
\centering
\resizebox{\linewidth}{!}{
\begin{tabular}{lccc}
\toprule
\multirow{2}{*}{\textbf{Methods}} &  \multicolumn{3}{c}{\textbf{RestBench-TMDB}} \\
\cmidrule(lr){2-4}
& Success & Accuracy & Path \\ 
\midrule
\rowcolor{mygray}
\textit{Qwen2.5-14B-Instruct}  &    &   &   \\
ReAct~\cite{DBLP:conf/iclr/YaoZYDSN023} & 67.0 & 40.0 & 42.0 \\ 
Chameleon~\cite{DBLP:conf/nips/LuPCGCWZG23} & 68.0 & 38.0 & 45.0 \\ 
CodeAct~\cite{DBLP:conf/icml/WangCY0L0J24}  & 72.0 & 58.0 & 62.5 \\
\texttt{ToolCoder} & \textbf{81.0} & \textbf{72.0} & \textbf{76.0} \\
\midrule
\rowcolor{mygray}
\textit{Qwen2.5-Coder-14B-Instruct}  &    &   &   \\
ReAct~\cite{DBLP:conf/iclr/YaoZYDSN023} & 68.0 & 41.5 & 47.0 \\ 
Chameleon~\cite{DBLP:conf/nips/LuPCGCWZG23} & 70.0 & 45.4 & 50.0 \\ 
CodeAct~\cite{DBLP:conf/icml/WangCY0L0J24}  & 80.0 & 65.0 & 75.0 \\
\texttt{ToolCoder} & \textbf{85.0} & \textbf{76.0} & \textbf{80.0} \\
\midrule
\rowcolor{mygray}
\textit{Qwen2.5-32B-Instruct}  &    &   &   \\
ReAct~\cite{DBLP:conf/iclr/YaoZYDSN023} & 73.0 & 50.0 & 51.0 \\ 
Chameleon~\cite{DBLP:conf/nips/LuPCGCWZG23} & 71.0 & 48.5 & 54.5 \\ 
CodeAct~\cite{DBLP:conf/icml/WangCY0L0J24}  & 82.0 & 62.5 & 76.5 \\
\texttt{ToolCoder} & \textbf{89.0} & \textbf{82.0} & \textbf{84.0} \\
\midrule
\rowcolor{mygray}
\textit{Qwen2.5-Coder-32B-Instruct}  &    &   &   \\
ReAct~\cite{DBLP:conf/iclr/YaoZYDSN023} & 76.5 & 56.5 & 58.0 \\ 
Chameleon~\cite{DBLP:conf/nips/LuPCGCWZG23} & 78.0 & 56.0 & 57.0 \\ 
CodeAct~\cite{DBLP:conf/icml/WangCY0L0J24}  & 92.0 & 82.0 & 85.0 \\
\texttt{ToolCoder} & \textbf{96.0} & \textbf{90.0} & \textbf{91.0} \\
\bottomrule
\end{tabular}}
\caption{Evaluate the performance of \texttt{ToolCoder} on different open source LLMs on RestBench-TMDB.}
\label{tab:opensource_lms}
\end{table}

\paragraph{Experiments on Open-Source LLMs}
To evaluate the effectiveness of \texttt{ToolCoder} with open-source LLMs, we conduct comprehensive experiments on RestBench-TMDB. The results in Table~\ref{tab:opensource_lms} reveal several significant findings. First, model scaling demonstrates consistent performance improvements across all methods. When scaling from 14B to 32B, we observe substantial gains in success rate, accuracy, and path correctness across all approaches. Second, the comparison between base models and their coder-enhanced counterparts reveals the substantial advantages of code-specialized capabilities. This is particularly evident in code-empowered methods like \texttt{ToolCoder} and CodeAct. These improvements demonstrate that code-specialized models are better equipped to handle the structured reasoning and precise execution required in tool learning tasks. Third, \texttt{ToolCoder} consistently outperforms all baselines across different model configurations. This superior performance validates the effectiveness of our code-empowered framework in leveraging the code-specific strengths of LLMs for tool learning.

\begin{table}[t]
\centering
\resizebox{0.8\linewidth}{!}{
\begin{tabular}{lcccc}
\toprule
\multirow{2}{*}{\textbf{Methods}} & \multicolumn{2}{c}{\textbf{RestBench}}  & \multicolumn{2}{c}{\textbf{API-Bank}}\\
\cline{2-3}\cline{4-5}
& TMDB & Spotify & LV1 & LV2 \\
\midrule
ReAct  & 7.78   & 8.68   & 6.45  & 8.38  \\
Chameleon   & 6.52    & 7.12    & 5.64  &  6.92   \\
ConAgents   &   6.78  &   7.68    &  5.45   &  7.38   \\
RestGPT   & 9.04   & 10.24   &  7.27  &   9.84   \\
\texttt{ToolCoder}    &  7.69  &  8.70   &  6.46  &  8.47 \\
\bottomrule
\end{tabular}}
\caption{Efficiency analysis of different methods on the RestBench and API-Bank datasets. We report the average number of API calls required per question, where a lower value indicates higher efficiency.}
\label{tab:efficiency}
\end{table}

\subsection{Efficiency Analysis}
To analyze the efficiency of API usage among different methods, we compare the average number of LLMs calls required per question on the RestBench and API-Bank benchmarks, as shown in Table~\ref{tab:efficiency}. \texttt{ToolCoder} demonstrates comparable efficiency to baseline methods such as ReAct, Chameleon, and ConAgents, with basically similar API usage. This result suggests that \texttt{ToolCoder} achieves higher correct path rates and success rates in task planning and execution compared to other methods, without incurring additional efficiency costs, thereby maintaining a practical balance between improved performance and resource consumption.


\section{Conclusion}
We present \texttt{ToolCoder}, a novel framework that reformulates tool learning as a code generation task, leveraging software engineering principles and Python function structures to enhance complex reasoning and execution reliability. By systematically transforming natural language queries into structured function scaffolds, incorporating descriptive comments for multi-step planning, and maintaining a repository of successfully executed functions, \texttt{ToolCoder} establishes a robust approach to tool learning. Its integrated error diagnosis mechanism further ensures adaptability based on code-based feedback. Experimental results demonstrate its superior performance in both task completion accuracy and execution reliability compared to existing methods, validating the effectiveness of our code-centric approach to tool learning.

\section*{Acknowledgement}
This work was supported by the Strategic Priority Research Program of the CAS under Grants No.XDB0680302, the National Natural Science Foundation of China (NSFC) under Grants No62276248, and the Youth Innovation Promotion Association CAS under Grants No. 2023111.

\section*{Limitations}
Despite the promising performance demonstrated by our approach, several limitations constrain its applicability and highlight directions for future improvement. First, the method relies heavily on the availability of clear, well-defined, and comprehensive API documentation. When such documentation is incomplete, ambiguous, or inconsistent, the model’s ability to infer tool behavior and execute tasks correctly is significantly hindered. This dependency limits the approach's robustness in real-world scenarios where documentation quality is not guaranteed. Second, our approach adopts a global planning strategy that excels in generating structured and efficient solutions but lacks the flexibility to address dynamic, real-time constraints in evolving environments. For example, unforeseen changes in task requirements or partial observability may require adaptive or incremental adjustments that our current framework does not fully support. Finally, scalability remains a challenge when dealing with tasks involving numerous tools and complex interactions. While our approach is effective for moderately complex tool planning, scenarios with a large number of interdependent tools may introduce combinatorial challenges, potentially resulting in suboptimal plans.


\bibliography{anthology,custom}

\appendix
\section{Appendix}
To provide an intuitive understanding of how our framework works, we present detailed examples in this section. We first provide example solutions for our \texttt{ToolCoder} for the query: \textit{give me the number of movies directed by Sofia Coppola}. All results are generated with \texttt{gpt-4o-mini}. Furthermore, we also display the prompts used to implement each module on the RestBench-TMDB dataset.

\begin{figure*}[htbp]
\centering
\begin{lstlisting}[language=python, basicstyle=\ttfamily\tiny]
def get_directed_movie_count(director_name: str) -> int:
    """
    Get the number of movies directed by a specific director.

    Parameters:
        director_name (str): The full name of the director whose movies to count.

    Returns:
        int: The number of movies directed by the specified director.
    """
    # Step 1. Search for the director "Sofia Coppola" using the /3/search/person API to find her person_id.
    # Step 2. Retrieve the movie credits for Sofia Coppola using her person_id with the /3/person/{person_id}/movie_credits API.
    
if __name__ == "__main__":
    number_of_movies_directed = get_directed_movie_count(director_name="Sofia Coppola")
    print("Number of movies directed by Sofia Coppola:", number_of_movies_directed)
\end{lstlisting}
\caption{An example of the task decomposition result $c_{\boldsymbol{s}}$ generated by \texttt{ToolCoder} for the query: \textit{give me the number of movies directed by Sofia Coppola}.}
\label{app:code_comment_example}
\end{figure*}
\begin{figure*}[htbp]
\centering
\begin{lstlisting}[language=python, basicstyle=\ttfamily\tiny]
def get_directed_movie_count(director_name: str) -> int:
    """
    Get the number of movies directed by a specific director.

    Parameters:
        director_name (str): The full name of the director whose movies to count.

    Returns:
        int: The number of movies directed by the specified director.
    """
    # Step 1. Search for the director "Sofia Coppola" using the /3/search/person API to find her person_id.
    person_search_response = call_api(
        api_path="/3/search/person", params={"query": director_name}
    )

    # Get person_id from response
    person_id = (
        person_search_response.get("results")[0]["id"]
        if person_search_response.get("results")
        else None
    )

    # Step 2. Retrieve the movie credits for Sofia Coppola using her person_id with the /3/person/{person_id}/movie_credits API.
    if person_id:
        movie_credits_response = call_api(
            api_path="/3/person/{person_id}/movie_credits",
            params={"person_id": person_id},
        )

        # Step 3. Count the number of movies in the credits retrieved.
        # Filter for movies that Sofia Coppola directed
        directed_movies = [
            movie
            for movie in movie_credits_response["crew"]
            if movie["job"] == "Director"
        ]

        number_of_movies_directed = len(directed_movies)
        return number_of_movies_directed

    return 0

if __name__ == "__main__":
    number_of_movies_directed = get_directed_movie_count(director_name="Sofia Coppola")
    print("Number of movies directed by Sofia Coppola:", number_of_movies_directed)
\end{lstlisting}
\caption{An example of the pseudo-code $c_{\boldsymbol{p}}$ generated by \texttt{ToolCoder} for the query: \textit{give me the number of movies directed by Sofia Coppola}.}
\label{app:planner_example}
\end{figure*}
\begin{figure*}[ht]
\centering
\tiny
\begin{lstlisting}[language=python, basicstyle=\ttfamily\tiny]
import requests

# Define the required headers for API access.
headers = {"Authorization": "YOUR API KEY"}

def search_person(query: str, include_adult=False, language="en-US", page=1):
    '''
    Omitted for simplicity due to space limitations
    '''
    url = "https://api.themoviedb.org/3/search/person"
    
    # Prepare the parameters for the request
    params = {
        "query": query,
        "include_adult": include_adult,
        "language": language,
        "page": page
    }
    
    # Make the API request
    response = requests.get(url, headers=headers, params=params)
    response.raise_for_status()
    return response.json()

def get_person_movie_credits(person_id: int, language="en-US"):
    '''
    Omitted for simplicity due to space limitations
    '''
    url = f"https://api.themoviedb.org/3/person/{person_id}/movie_credits"
    
    # Prepare query parameters
    params = {"language": language}
    
    # Make the API request
    response = requests.get(url, headers=headers, params=params)
    response.raise_for_status()
    return response.json()

def get_directed_movie_count(director_name: str) -> int:
    '''
    Get the number of movies directed by a specific director.

    Parameters:
        director_name (str): The full name of the director whose movies to count.

    Returns:
        int: The number of movies directed by the specified director.
    '''
    # Step 1. Search for the director "Sofia Coppola" using the /3/search/person API to find her person_id.
    person_search_response = search_person(query=director_name)

    # Get person_id from response
    person_id = (
        person_search_response.get("results")[0]["id"]
        if person_search_response and person_search_response.get("results")
        else None
    )

    # Step 2. Retrieve the movie credits for Sofia Coppola using her person_id with the /3/person/{person_id}/movie_credits API.
    if person_id:
        movie_credits_response = get_person_movie_credits(person_id)

        # Step 3. Count the number of movies in the credits retrieved.
        # Filter for movies that Sofia Coppola directed
        directed_movies = [
            movie
            for movie in movie_credits_response["crew"]
            if movie["job"] == "Director"
        ]

        number_of_movies_directed = len(directed_movies)
        return number_of_movies_directed

    return 0

if __name__ == "__main__":
    number_of_movies_directed = get_directed_movie_count(director_name="Sofia Coppola")
    print("Number of movies directed by Sofia Coppola:", number_of_movies_directed)
\end{lstlisting}
\caption{An code example of the main function $F$ generated by \texttt{ToolCoder} for the query: \textit{give me the number of movies directed by Sofia Coppola}. This step implements the actual functionality of all API calls in the \texttt{call\_api} placeholders in Figure~\ref{app:planner_example} according to their API documentations. Due to space limitations, the docstrings of some functions have been omitted.}
\label{app:main_function_example}
\end{figure*}

\begin{figure*}[ht]
\centering
\begin{tcolorbox}[colback=white!95!blue, colframe=blue!40!, title=Prompt for Task to Code Module $\mathcal{M}_{T2C}$, width=\textwidth]
\small
\texttt{You are a coding assistant specialized in converting task requirements into precise Python function definitions. Your role is to define a function that fully meets the task's objectives by specifying a clear function name, parameters, return type, and docstring.}\\
\\
\texttt{\#\#\# Instructions:}\\
\texttt{For each given task, carefully follow these steps to create a well-structured function definition:}\\
\texttt{1. Function Name: Choose a concise, meaningful function name that accurately reflects the task's purpose.}\\
\texttt{2. Parameters: Identify and define the essential input(s) the function requires. For each parameter:}\\
\texttt{\parbox{2pt}{\hspace{8em}} - Use a descriptive name and include a type annotation.}\\
\texttt{3. Return Type: Specify an appropriate return type (e.g., 'int', 'str', 'dict', 'list') that best represents the function’s output.}\\
\texttt{4. Docstring: Write a detailed, clear docstring that includes:}\\
\texttt{\parbox{2pt}{\hspace{8em}} - A brief description of the function’s purpose.}\\
\texttt{\parbox{2pt}{\hspace{8em}} - Descriptions for each parameter, detailing its type and role.}\\
\texttt{\parbox{2pt}{\hspace{8em}} - Explanation of the return value, with type and a brief on what it represents.}\\
\texttt{\parbox{2pt}{\hspace{8em}} - Explicit Notes on using 'call\_api', with the following rules:}\\
\texttt{\parbox{2pt}{\hspace{8em}} \quad - Each 'call\_api' invocation must use an 'api\_path' from the provided toolbox; do not invent or alter paths.}\\
\texttt{\parbox{2pt}{\hspace{8em}} \quad - Place all request parameters within the 'params' dictionary; do not format 'api\_path' using string interpolation or formatted strings.}\\
\texttt{\parbox{2pt}{\hspace{8em}} \quad - Example: use 'call\_api(api\_path="\//3\//movie\//\{movie\_id\}\//credits", params=\{"movie\_id": movie\_id\})' instead of 'call\_api(api\_path=f"\//3\//movie\//{movie\_id}\//credits", params=\{\})'.}\\
\texttt{\parbox{2pt}{\hspace{8em}} \quad - Refer to the example functions below for guidance on structuring this.}\\
\texttt{\parbox{2pt}{\hspace{8em}}  - Use triple single quotes for the docstring, as shown below.}\\
\texttt{5. Function Body: Leave the function body empty. Only include the function definition, parameters, and docstring in your output.}\\

\texttt{Now, let's begin!}\\
\texttt{You are given a question: \{question\}}\\
\texttt{Python Function:}
\end{tcolorbox}
\caption{The prompt used for implementing the task-to-code function module $\mathcal{M}_{T2C}$ in the RestBench-TMDB dataset during our experiments.}
\label{fig:task_to_code_prompt}
\end{figure*}
\begin{figure*}[ht]
\centering
\begin{tcolorbox}[colback=white!95!blue, colframe=blue!40!, title=Prompt for Subtask Planning Module $\mathcal{M}_{TP}$, width=\textwidth]
\small
\texttt{Your objective is to analyze the user's complex question and create a breakdown of actionable subtasks, identifying the most appropriate APIs from the provided toolbox to address each subtask in sequence.}\\

\texttt{\#\#\# Available Tools:} \\
\texttt{\{toolbox\}}\\

\texttt{Each API includes a request path and a description of its functionality. Utilize these APIs strategically to decompose and solve the user's question according to the following guidelines:}\\

\texttt{\#\#\# Guidelines}\\
\texttt{1. Clarify Requirements: Carefully read the user's question to determine key requirements, objectives, and expected outcomes. Identify any specific data or information you need to gather to satisfy the request.}\\
\texttt{2. Break Down the Task: Divide the complex task into clear, manageable subtasks that each address part of the user's needs and can be fulfilled using one or more APIs.}\\
\texttt{3. Select Relevant APIs: Match each subtask to the most relevant API(s) from the toolbox. Base your choices on each API's functionality and response format to ensure it provides the required information.}\\
\texttt{4. Handle Dependencies: Check for dependencies between APIs. For example, to access '\//3\//movie\//\{movie\_id\}\//credits', the 'movie\_id' must first be retrieved via a compatible API.}\\
\texttt{5. Output in Pseudo-Code: Write your solution as a Python function using pseudo-code, with each step annotated in comments (denote as Step 1 to N) describing the subtask. Do not write any code implementation for the steps—only document the subtasks in comments.}\\

\texttt{User's Question: \{question\}}\\
\texttt{Pseudo-Code Task: \{pseudo\_code\_task\}}\\
\texttt{Planned Subtasks:}
\end{tcolorbox}
\caption{The prompt used for implementing the subtask planning module $\mathcal{M}_{TP}$ in the RestBench-TMDB dataset during our experiments.}
\label{fig:subtask_planning_prompt}
\end{figure*}
\begin{figure*}[ht]
\centering
\begin{tcolorbox}[colback=white!95!blue, colframe=blue!40!, title=Prompt for Tool Selection Module $\mathcal{M}_{TS}$, width=\textwidth]
\small
\texttt{Your task is to help the user select appropriate APIs from the provided toolbox and complete the function template to solve the problem accurately.}\\

\texttt{\#\#\# Instructions}\\
\texttt{1. Input Provided:}\\
\parbox{2pt}{\hspace{8em}} \texttt{- Problem Description: A specific question or request from the user (e.g., "give me the number of movies directed by Sofia Coppola").}\\
\parbox{2pt}{\hspace{8em}} \texttt{- Pseudo-Code Task Template: A function template with the following components:}\\
\parbox{2pt}{\hspace{8em}} \quad \texttt{- Function signature, parameters, return type, and a functional description.}\\
\parbox{2pt}{\hspace{8em}} \quad \texttt{- Structured subtask comments directly embedded in the pseudo-code template, outlining each planned subtask step-by-step.}\\

\texttt{2. Expected Output:}\\
\parbox{2pt}{\hspace{8em}} \texttt{- Based on the problem description and the structured action plan provided, complete the pseudo-code by selecting and integrating suitable API calls from the provided candidate toolbox.}\\
\parbox{2pt}{\hspace{8em}} \texttt{- Use the placeholder 'call\_api(api\_path, params)' for each API call, where:}\\
\parbox{2pt}{\hspace{8em}} \quad \texttt{- 'api\_path' strictly matches a valid, existing API path from the provided candidate toolbox. Do not create or assume any non-existent API paths. No 'api\_path' outside this toolbox should be used or inferred.}\\
\parbox{2pt}{\hspace{8em}} \quad \texttt{- Place all request parameters within the 'params' dictionary; do not format 'api\_path' using string interpolation or formatted strings.}\\
\parbox{2pt}{\hspace{8em}} \quad \texttt{- Example: use 'call\_api(api\_path="\//3\//movie\//\{movie\_id\}\//credits", params=\{"movie\_id": movie\_id\})' instead of 'call\_api(api\_path=f"\//3\//movie\//{movie\_id}\//credits", params=\{\})'.}\\

\texttt{3. Guidelines for API Selection:}\\
\parbox{2pt}{\hspace{8em}} \texttt{- Do not provide a direct answer to the problem. Instead, fill in the pseudo-code template to enable successful execution.}\\
\parbox{2pt}{\hspace{8em}} \texttt{- Follow these steps for each 'call\_api' integration:}\\
\parbox{2pt}{\hspace{8em}} \quad \texttt{- Identify any necessary preliminary calls (e.g., fetching an entity ID).}\\
\parbox{2pt}{\hspace{8em}} \quad \texttt{- Ensure each step logically contributes to the function’s purpose, using helper variables and conditional checks where needed.}\\
\parbox{2pt}{\hspace{8em}} \quad \texttt{- Keep pseudo-code succinct, adding comments that clarify the role of each API call.}\\

\texttt{\#\#\# Provided API toolbox:}\\
\texttt{\{toolbox\}}\\

\texttt{Let's begin!}\\
\texttt{Question: \{question\}}\\
\texttt{Pseudo-Code Task: \{pseudo\_code\_task\}}\\
\texttt{Please complete the pseudo-code solution:}
\end{tcolorbox}
\caption{The prompt used for implementing the tool selection module $\mathcal{M}_{TS}$ in the RestBench-TMDB dataset during our experiments.}
\label{fig:tool_selection_prompt}
\end{figure*}
\begin{figure*}[ht]
\centering
\begin{tcolorbox}[colback=white!95!blue, colframe=blue!40!, title=Prompt for Code Generation Module $\mathcal{M}_{CG}$, width=\textwidth]
\small
\texttt{You are a coding assistant specializing in Python and API integration. Your task is to translate code snippets with 'call\_api' placeholders into executable Python code. Each 'call\_api' placeholder represents an API call that should be implemented as a separate function.}\\

\texttt{\#\#\# Input:}\\
\texttt{1. A Python code snippet containing one or more 'call\_api' placeholders.}\\
\texttt{2. The OpenAPI documentation describing the APIs, including:}\\
\texttt{\parbox{2pt}{\hspace{8em}} - Endpoints ('path'),}\\
\texttt{\parbox{2pt}{\hspace{8em}} - Expected input parameters ('parameters'),}\\
\texttt{\parbox{2pt}{\hspace{8em}} - Response structure ('schema').}\\

\texttt{\#\#\# Requirements:}\\
\texttt{1. Function Creation: For each API in the OpenAPI specification:}\\
\texttt{\parbox{2pt}{\hspace{8em}} - First, check if the API's dictionary in the documentation contains a 'reusable\_code' field. If present:}\\
\texttt{\parbox{2pt}{\hspace{8em}} \quad - Extract the function code from the 'reusable\_code' field and reuse it to implement the corresponding API functionality in 'call\_api'.}\\
\texttt{\parbox{2pt}{\hspace{8em}} \quad - Make minimal modifications if necessary to integrate the function with the current codebase, while preserving its original logic.}\\
\texttt{\parbox{2pt}{\hspace{8em}} - If the 'reusable\_code' field is not present:}\\
\texttt{\parbox{2pt}{\hspace{8em}} \quad - Implement a Python function using the 'requests' library.}\\
\texttt{\parbox{2pt}{\hspace{8em}} \quad - Use 'https://api.themoviedb.org' as the 'base\_url'. Construct the full URL by appending the 'api\_path' to this 'base\_url'.}\\
\texttt{\parbox{2pt}{\hspace{8em}} \quad - Use the 'GET' method for all requests.}\\
\texttt{\parbox{2pt}{\hspace{8em}} \quad - Include two parameters in the function signature:}\\
\texttt{\parbox{2pt}{\hspace{8em}} \quad \quad - 'params': An optional dictionary for query parameters.}\\
\texttt{\parbox{2pt}{\hspace{8em}} \quad \quad - 'headers': A mandatory dictionary containing the following: headers = \{"Authorization": "Bearer \{API\_KEY\}"\}}\\
\texttt{\parbox{2pt}{\hspace{8em}} \quad \quad - Ensure the function returns parsed response data in a usable format.}\\
\texttt{2. Integration: Replace each 'call\_api' placeholder in the original code with calls to the corresponding functions, ensuring that the replacement maintains the original logic.}\\
\texttt{3. Error Handling: Add error handling to manage API failures (e.g., HTTP errors). Use 'try/except' blocks to log errors or raise exceptions.}\\
\texttt{4. Code Quality:}\\
\texttt{\parbox{2pt}{\hspace{8em}}- Write clear and descriptive docstrings for each function.}\\
\texttt{\parbox{2pt}{\hspace{8em}}- Follow Python best practices for readability and maintainability.}\\

\texttt{\#\#\# Additional Notes:}\\
\texttt{\parbox{2pt}{\hspace{8em}} - Validate input parameters where applicable.}\\
\texttt{\parbox{2pt}{\hspace{8em}} - Ensure all requests include the 'headers' defined above.}\\
\texttt{\parbox{2pt}{\hspace{8em}} - Clearly document any assumptions made during implementation.}\\
\texttt{\parbox{2pt}{\hspace{8em}} - If reusing functions from the 'reusable\_code' field, ensure their integration complies with Python conventions and the overall project architecture.}\\

\texttt{\#\#\# Output:}\\
\texttt{The complete, executable Python code with:}\\
\texttt{\parbox{2pt}{\hspace{8em}} - Defined functions for each API,}\\
\texttt{\parbox{2pt}{\hspace{8em}} - Updated logic that calls these functions.}\\

\texttt{Now, it's your turn!}\\
\texttt{\#\#\# Question: \{question\}}\\
\texttt{\#\#\# Input:
\texttt{\{code\_solution\}}}\\
\texttt{\#\#\# OpenAPI Documents as well as a reusable code snippet (optional):}\\
\texttt{\{api\_doc\}}\\
\texttt{\#\#\# Output:}\\
\end{tcolorbox}
\caption{The prompt utilized for implementing the code generation module $\mathcal{M}_{CG}$ in the RestBench-TMDB dataset for our experiments.}
\label{fig:main_function_template}
\end{figure*}

\end{document}